\documentclass[review,3p,times]{elsarticle}

\usepackage{lmodern} 
\usepackage{amsmath,amssymb,amsthm}
\usepackage{algorithm,algpseudocode}
\usepackage{graphicx}
\usepackage{booktabs}
\usepackage{multirow}
\usepackage{xcolor}
\usepackage{hyperref}
\usepackage{url}
\usepackage{natbib}
\usepackage{lineno}
\usepackage{tikz}
\usetikzlibrary{positioning,arrows.meta,fit,backgrounds}

\hypersetup{
    colorlinks=true,
    linkcolor=blue!50!black,
    citecolor=blue!50!black,
    urlcolor=blue!50!black,
}

\newcommand{\method}{TypedCSIP}
\newcommand{\dataset}{LCR-CN}
\newcommand{\vone}{\method{}-v1}
\newcommand{\vtwo}{\method{}-v2}
\newcommand{\superior}{A}
\newcommand{\subordinate}{B}
\newcommand{\golden}{\widetilde{B}}
\newcommand{\Ltype}{\mathcal{L}_{\mathrm{type}}}
\newcommand{\Lselect}{\mathcal{L}_{\mathrm{select}}}

\journal{Information Processing \& Management}

\begin{document}


\begin{frontmatter}

\title{
  \method{}: Typed Counterfactual Pretraining \\
  for Chinese Legislative Conflict Classification
}

\author[1,2]{Yao Liu\corref{cor1}}
\ead{liuyao@student.usm.my}
\author[2]{Tien Ping Tan}

\address[1]{Chengdu University of Technology, Leshan, China}
\address[2]{School of Computer Sciences, Universiti Sains Malaysia, Penang, Malaysia}

\cortext[cor1]{Corresponding author.}

\begin{abstract}
We introduce \method{}, a typed counterfactual pretraining method for
the conflict-classification task of the \dataset{} benchmark
\citep{zhao2026lcrcn}: given a $(\superior, \subordinate)$ provision
pair, predict whether they conflict and which of four legal-doctrine
types describes the inconsistency. The method exploits \dataset{}'s
expert-written minimal revisions $\golden$ --- a downstream correction
annotation released alongside the classification labels --- as
training-time counterfactual supervision; at test time the classifier
reads only $(\superior, \subordinate)$. \method{} therefore applies to
benchmarks that ship expert minimal-revision annotations alongside
classification labels; we make no claim of applicability to
classification benchmarks that do not. We registered the confirmatory
test on the Open Science Framework before observing any v6
measurements: 18 PRNG-derived seeds and a locked seed-as-unit decision
rule requiring the mean per-seed difference to clear $0.8$~pp and both
the seed-bootstrap and Student-$t$ 95\% lower bounds to clear zero. On
the official 696-record test split, the pre-registered \vtwo{} variant
improves macro-F$_1$ over the strongest single-model baseline of the
dataset paper by $+0.916$~pp on the primary
\texttt{chinese-roberta-wwm-ext} backbone and by $+1.288$~pp on the
cross-backbone \texttt{SAILER} replication; both cells pass the locked
rule. Stage~1 pretrains a shared encoder with a typed Counterfactual
Selective Intervention Pretraining objective on
$(\superior, \subordinate, \golden)$ triplets, treating $\golden$ as a
counterfactual that the typed factor head must classify as carrying no
conflict evidence. Stage~2 then transfers the encoder to the five-way
classification head. A matched-seed
auxiliary on the simpler \vone{} variant (typed-discard transfer,
without replay) reaches mean $\Delta = +1.137$~pp on RoBERTa, clearing
the descriptive C1$'$ label that the pre-registered \vtwo{} cell does
not reach on this backbone; the anti-forget replay redesign is
therefore not load-bearing for the family-level gain. Because the
official split has structural $(\superior, \golden)$ tuple overlap
by design (452 of 696 test records share a tuple with training;
Section~\ref{sec:task-data:audit}), we additionally report a cold-start
stratified result on the 244 Unseen-gB test records (those whose
$(\superior, \golden)$ tuple is novel relative to training). The
\vtwo{} gain stays positive with both confidence intervals above zero
on both backbones, and concentrates in the conflict-type classes
rather than the no-conflict class. The Stage-2 encoder is
classification-specialized: a cross-task diagnostic that reuses it as a
single-vector bi-encoder for \dataset{} Task~1 (superior-law retrieval)
gives Accuracy@1 $-20.83$~pp, so we scope the contribution to the
conflict-classification task and treat retrieval and revision
generation as outside the present study. We release all code, all $72$
pre-registered test-prediction files (18 seeds $\times$ 2 cells
$\times$ 2 backbones), the $18$ matched-seed \vone{} auxiliary files,
the $18$ MLM-continuation control files registered as addendum~2, and
the OSF pre-registration record.
\end{abstract}


\end{frontmatter}



\section{Introduction}
\label{sec:intro}

Subordinate legislation must stay consistent with the superior law
above it, and verifying that consistency is a recurring,
labor-intensive task in legal practice. Jurisdictions address it in
different ways: some through judicial review for unconstitutionality or
ultra vires, others through dedicated review bodies. China handles it
through a centralized, high-volume filing-and-review system, in which
thousands of subordinate provisions are examined by hand each year,
each checked against superior law to locate the point of conflict and
draft a minimal corrective revision, so that its four-tier hierarchy of
constitution, national law, administrative regulation, and local rule
stays internally consistent.

Legal NLP is broadly active: there is substantial work on judgment
prediction \citep{xiao2018cail}, legal case retrieval
\citep{ma2021lecard,xiao2021lawformer,li2023sailer,deng2024keller},
and LLM-scale legal-reasoning benchmarks \citep{guha2023legalbench}.
Legislative conflict classification, however, is almost untouched.
The closest international work, Clash-of-Leges
\citep{italiani2026clash}, performs only a binary conflict/no-conflict
judgment and targets Italian law; before \dataset{}
\citep{zhao2026lcrcn}, no work had systematically classified conflicts
between Chinese superior and subordinate legislation into specific
types. The task is also not easy: the evidence that fixes which type a
conflict belongs to is often a small, localized span within a long
pair of provisions, and a few changed words can move a conflict from
one type to another. Existing methods compress the whole decision into
a single class label and train a classifier against it
\citep{lamgfm2023,gjudge2024}, as do the \dataset{} baselines
\citep{zhao2026lcrcn}; a model trained this way sees only the final
verdict and cannot locate the decisive evidence.

We propose \method{}, whose starting point is an information source
that existing methods leave unused: the expert revision. Every
conflicting subordinate provision comes with an expert-written
revision that removes the conflict and leaves the rest of the text
word for word unchanged. This revision pinpoints the conflict on its
own: where the expert edits marks where the conflict lies, and how
they edit reveals which type it is. \method{} lets the model learn
directly from these expert revisions, recognizing a conflict from how
an expert resolves it.

Concretely, \method{} proceeds in two steps. The first step is
pretraining: the model learns from the contrast between a conflicting
provision and its expert revision, building the ability to recognize
conflicts. The second step is fine-tuning: on top of this model, a
five-way classifier is trained to label a new subordinate provision as
one of the four conflict types, or as no conflict.

The entire confirmatory analysis (cells, backbones, seeds, training
scripts, analysis script, and locked decision rule) is registered on
the Open Science Framework (OSF) before any v6 measurement is
observed; the 18 primary seeds derive deterministically from an
earlier pre-registration's SHA-256 hash, removing post-hoc seed
selection. Under the locked seed-as-unit rule (mean per-seed
difference $\geq 0.8$~pp and both 95\% lower bounds above zero), the
pre-registered \vtwo{} cell passes on the primary
\texttt{chinese-roberta-wwm-ext} backbone (mean $\Delta = +0.916$~pp)
and on the cross-backbone \texttt{SAILER} replication (mean $\Delta =
+1.288$~pp).

A matched-seed auxiliary on the simpler \vone{} variant reaches mean
$\Delta = +1.137$~pp on RoBERTa, clearing the descriptive C1$'$ label
($\geq 1.0$~pp) that the pre-registered \vtwo{} cell does not reach on
this backbone. The anti-forget replay redesign is therefore not
load-bearing for the family-level gain. We interpret the load-bearing
component of the family as the Stage-1 typed CSIP signal that both
variants inherit, with the simpler typed-discard transfer of \vone{}
sufficient to surface it. A cold-start stratified evaluation that
restricts the test set to items whose $(\superior, \golden)$ tuple
does not appear in training leaves the \vtwo{} gain positive with
both confidence intervals above zero on both backbones; a per-class
breakdown of the Unseen-gB stratum (Section~\ref{sec:audits:strat})
shows the gain is concentrated in the conflict-type classes
(Definition $+8.7$~pp on RoBERTa, Condition $+3.2$~pp on both
backbones) rather than the No-Conflict class, which is inconsistent
with an account on which the cold-start gain is driven by NC-class
composition shift.

The remainder of the paper details the typed CSIP objective and the
two Stage-2 variants (Section~\ref{sec:method}), the pre-registration
protocol (Section~\ref{sec:prereg}), the confirmatory results on both
backbones (Section~\ref{sec:experiments}), the cold-start and
matched-seed audits (Section~\ref{sec:audits}), and an interpretive
discussion that frames the typed-discard transfer as the load-bearing
component of the family (Section~\ref{sec:discussion}).


\section{Related Work}
\label{sec:related}

\subsection{Chinese legal natural language processing}
\label{sec:related:chinese-legal}

Chinese legal NLP has accumulated a body of domain-adapted encoders and
benchmark tasks. \citet{xiao2021lawformer} pretrain a long-document
encoder for Chinese legal texts. \citet{li2023sailer} introduce SAILER, a
structure-aware Chinese legal encoder optimized for case retrieval, which
we use as the cross-backbone replication target in this paper.
\citet{deng2024keller} extend the legal information retrieval (IR) line with knowledge-enhanced
ranking; KELLER reports state-of-the-art numbers on LeCaRD-style
benchmarks but does not address conflict classification. Legal judgment
prediction is the most heavily studied downstream task: in this venue
alone, \citet{gjudge2024} propose graph-boosted constraints for CAIL
multi-task LJP, and \citet{lamgfm2023} introduce sememe-enhanced graph
fusion to discriminate confusable charges. Both target the CAIL family
rather than the conflict-review task we address here. Outside the legal
domain, \citet{sgbert2025} present a two-stage Chinese contrastive
pretraining method (SGBERT) that injects character-morphology knowledge;
the two-stage curriculum design parallels our own pretrain-then-transfer
recipe, but the supervision signal and downstream task differ.

Internationally, the closest precedent is Clash-of-Leges
\citep{italiani2026clash}, a bilingual (Italian/English) dataset of
conflicts between legal articles, derived from rulings of the Italian
Constitutional Court. Its conflict task is binary (conflict / no
conflict): it neither sorts conflicts into specific types nor targets
the hierarchical relation between superior and subordinate law. In the
Chinese setting, the \dataset{} benchmark of \citet{zhao2026lcrcn} is
the only public resource for legislative conflict review; it proposes
three benchmark tasks and reports fine-tuned encoder baselines (BERT,
RoBERTa, BERT-WWM, ERNIE) and a Qwen-LoRA baseline. Its
conflict-classification task has since had no dedicated method work.

\subsection{Legal text classification beyond judgment prediction}
\label{sec:related:other-legal-tasks}

The dominant downstream tasks in Chinese legal NLP are legal judgment
prediction (LJP) and legal case retrieval. LJP is benchmarked through
the CAIL family of datasets \citep{xiao2018cail}, on which the
graph-boosted constraint method of \citet{gjudge2024} and the
sememe-enhanced fusion of \citet{lamgfm2023} are recent representative
contributions in this venue. Legal case retrieval is benchmarked
through LeCaRD \citep{ma2021lecard}; SAILER \citep{li2023sailer} and
KELLER \citep{deng2024keller} are the strongest dense rankers on those
splits, and KELLER reports leading results on the LeCaRD~v2 extension.
LegalBench \citep{guha2023legalbench} extends the picture to
English-language legal reasoning at LLM scale across more than
160 tasks. These three task families share a single-document
or query--document framing where the classification or ranking decision
is conditioned on one input span. Conflict review is structurally
different: the unit of decision is a \emph{pair} of legal provisions
drawn from different layers of the regulatory hierarchy, the supervision
signal includes an expert-written revision of one element of the pair,
and the label space encodes legal-doctrine categories rather than a
charge or a relevance grade. The \dataset{} release isolates this
specific decision step, and method work developed on the adjacent
benchmarks above does not transfer to it directly.

\subsection{Contrastive and counterfactual pretraining}
\label{sec:related:contrastive}

Contrastive learning is a prominent pretraining recipe for
sentence-level representation \citep{gao2021simcse}; the
pretrain-then-discard-projection-head convention follows from the
original SimCLR work \citep{chen2020simclr}. For text classification
specifically, \citet{rcl2024} adapt the contrastive recipe with
retrieval-augmented positives for aspect sentiment, and
\citet{triplecl2024} extend the supervised contrastive objective to
label-aware multiclass settings; both report SOTA on their respective
benchmarks but neither addresses counterfactual supervision or the
two-stage transfer pattern we study.

The counterfactual contrastive line traces to \citet{kaushik2020cad},
who introduced counterfactually-augmented data by having human annotators
make minimal revisions to documents that flip the gold label, then
training models that use these paired counterfactuals as supervision.
Our expert revision $\golden$ inherits this primitive: a minimal
expert-written modification of one element of the input pair that
resolves the conflict labeled at training time. Two more recent
methodological neighbors share the paired-counterfactual structure.
\citet{roschewitz2024cfsimclr} introduce CF-SimCLR for medical image
representation learning, using causally synthesized images as
counterfactual positives. \citet{qiu2024paircfr} apply a paired-CAD
plus contrastive-loss recipe to sentiment analysis and natural language
inference, adding a contrastive term so the model uses global features
rather than only the edited tokens. Our pretraining objective shares
structure with both --- a counterfactual variant of the input contrasted
with the original under a contrastive loss --- but differs in three
ways: (i) our counterfactual $\golden$ is an expert-written legal
revision rather than a synthesized image or sentiment-flipping edit,
(ii) we attach a task-typed factor head whose denominator is the legal
conflict taxonomy rather than a generic projection MLP or single
contrastive logit, and (iii) we then transfer the pretrained encoder
to a fine-tuning task that uses the same five-way label space as the
pretraining factor head, allowing the typed-discard transfer pattern
to operate. In the NLP setting, \citet{ccprefix2024} apply
counterfactual contrastive prefix-tuning to many-class classification;
their counterfactuals are prefix-token perturbations, not domain-expert
revisions, and they target verbalizer ambiguity rather than
rule-violation classification.

Our work also relates to anti-forgetting techniques from continual
learning \citep{kirkpatrick2017ewc, lwf2018}. The \vtwo{} variant's
Stage-2 replay term is a domain-specific instance of generic rehearsal:
the Stage-1 pretraining loss is re-applied during Stage-2 fine-tuning
with a fixed coefficient. The matched-seed audit in
Section~\ref{sec:audits:v1} shows that this anti-forgetting term is not
load-bearing for the family-level gain on our task; the simpler
typed-discard \vone{} achieves a comparable mean delta.

\subsection{Reproducibility and pre-registration in empirical NLP}
\label{sec:related:prereg}

Pre-registration --- the practice of committing the experimental
protocol and analysis plan to a third-party time-stamped record before
collecting the data --- has been recommended in psychology and biomedical
research for over a decade \citep{nosek2018preregistration,
simmons2011falsepositive}. Within NLP the practice remains uncommon.
\citet{qiu2024cmcl} explicitly anchor an experimental procedure
on the Open Science Framework as part of a CMCL 2024 study, demonstrating
that the discipline is feasible in NLP venues. We follow this precedent:
our v6 confirmatory design (locked decision rule, fixed-$N$, PRNG-derived
seeds, no-extension commitment) was registered on OSF before any v6
measurement existed, and the registered file's SHA-256 fingerprint
anchors the seeds and the analysis script. We do not claim primacy for
OSF anchoring in NLP; we apply it because the locked rule discipline
removes seed selection and analysis-script choice as degrees of freedom
in the sense of \citet{steegen2016multiverse}.

A separate body of work studies reproducibility of NLP results
empirically (e.g., reproducibility studies of fine-tuned language models
across seeds and hyperparameters). Our pre-registered fixed-$N{=}18$
protocol with a locked seed-as-unit decision rule sits one level above
reproducibility-of-numbers: we commit in advance to the analysis that
will be applied to the seed-level observations, removing post-hoc
estimator selection as a source of inflation. We adopt paired-bootstrap
tools common in NLP system comparison \citep{koehn2004bootstrap}, lifted
from the example-level setting to seed-as-unit.

\subsection{Positioning of our contribution}
\label{sec:related:position}

The \method{} family combines three threads of prior work: the typed
Chinese legal encoder thread (Lawformer, SAILER), the counterfactual
contrastive pretraining thread (CF-SimCLR, Co$^2$PT), and the
SimCLR-style pretrain-then-discard transfer thread. The novel
combination is (a) the typed factor head whose denominator matches the
downstream taxonomy, (b) the use of expert-written legal revisions as
counterfactual positives, and (c) the typed-discard transfer pattern that
re-initializes the classification head while inheriting the pretrained
encoder. To these methodological elements we add (d) the
pre-registered confirmatory protocol described in Section~\ref{sec:prereg}
and (e) the cold-start stratified evaluation along the
$(\superior, \golden)$-tuple novelty axis
(Section~\ref{sec:audits:strat}). None of (a)--(c) is
individually novel in isolation; the combination on the conflict-review
task is new.


\section{Task and Data}
\label{sec:task-data}

\subsection{Task: Chinese legislative conflict classification}
\label{sec:task-data:task}

The Chinese Legal Conflict Review benchmark \dataset{} \citep{zhao2026lcrcn}
defines an automated taxonomic classification task over pairs of legal
provisions drawn from the Chinese legislative hierarchy. The input is a
pair $(\superior, \subordinate)$, where $\superior$ is a superior law
provision (national statute or higher-level regulation) and
$\subordinate$ is a subordinate provision under review. The output is a
single label drawn from five mutually exclusive categories: Responsibility
conflict, Condition conflict, Sanction conflict, Definition conflict, or
No-Conflict. Each conflicting pair is additionally accompanied by an
expert-written minimal revision $\golden$ of $\subordinate$ that
eliminates the conflict while preserving the regulatory intent. This
$\golden$ revision is annotated by legal-domain experts during dataset
construction and is provided as a separate field in the dataset rather
than as an explicit input to the classification task.

\dataset{} is the first publicly released benchmark for Chinese
legislative conflict review. The dataset paper of \citet{zhao2026lcrcn}
introduces three benchmark tasks: Superior Law Retrieval, Conflict
Classification, and Conflict Explanation and Revision Generation. The
present paper addresses only the second task; the first is an upstream
information retrieval problem and the third is a downstream generation
problem, both of which lie outside our scope. To our knowledge no prior
method paper has targeted the conflict classification head of \dataset{}.

\subsection{Data structure and splits}
\label{sec:task-data:splits}

The released benchmark contains a single fixed train/validation/test
partition with the canonical sizes 4895 / 1404 / 696 reported by
\citet{zhao2026lcrcn}. Our data ingestion pipeline applies two filters
in fixed order to the raw release: (i) records that fail strict JSON
decoding are skipped; (ii) records whose \texttt{conflict\_type\_en}
field does not map to one of the five canonical labels are skipped.
After both filters, $4{,}828$ training records enter the training pool;
the validation and test splits load cleanly at their nominal sizes
($1{,}404$ and $696$ records, respectively).
The pipeline does not perform within-split deduplication; any
near-duplicate subordinate provisions present in the released training
file are retained.

For Stage-1 CSIP pretraining (Section~\ref{sec:method:stage1}), we
apply two additional filters \emph{only to the CSIP-triplet loader}
without altering the Stage-2 fine-tuning pool: conflicting records
with an empty $\golden$ field are dropped, and conflicting records
whose $\golden$ string is byte-identical to the conflicting
$\subordinate$ are dropped, because the squared-difference selectivity
term in Eq.~\ref{eq:l-select} would be trivially zero in those cases.
The Stage-2 fine-tuning loader uses the full $4{,}828$-record training
pool regardless of whether $\golden$ is usable.

Each record contains, at minimum, an identifier, the superior provision
text, the subordinate provision text, the conflict-type label, and for
conflicting records the expert revision $\golden$. Additional metadata
fields (provision URL, document title, high-level-laws array) are present
but unused in our experiments. The five-way label distribution on the
held-out test split is reported in Table~\ref{tab:class-counts}; the
Definition class is the rarest at $n{=}48$, which informs our use of
class-weighted cross-entropy at Stage~2.

\begin{table}[t]
  \centering
  \small
  \begin{tabular}{lrrr}
    \toprule
    Class & Train (n) & Val (n) & Test (n) \\
    \midrule
    Responsibility       & 1617 & 464 & 229 \\
    Condition            & 1152 & 330 & 164 \\
    Sanction             &  935 & 268 & 133 \\
    Definition           &  333 &  96 &  48 \\
    No-Conflict          &  791 & 246 & 122 \\
    \midrule
    Total                & 4828 & 1404 & 696 \\
    \bottomrule
  \end{tabular}
  \caption{\dataset{} class distribution per split after our ingest
  filters (Section~\ref{sec:task-data:splits}). The Definition class is
  the rarest at every split ($n{=}48$ in test); our Stage~2 fine-tuning
  uses inverse-frequency class weights to address the imbalance. Test
  per-class counts match those reported by \citet{zhao2026lcrcn};
  train per-class counts are derived from the inverse-frequency weight
  vector logged at training time.}
  \label{tab:class-counts}
\end{table}

\subsection{$(\superior, \golden)$ tuple overlap and stratification basis}
\label{sec:task-data:audit}

The pretraining objective in Section~\ref{sec:method:stage1} uses the
expert revision $\golden$ as a counterfactual. To define a stratified
evaluation that complements the aggregate macro-F$_1$ on the official
test split, we characterize the cross-split structure of the
$(\superior, \golden)$ tuple, computing MD5 hashes of the joined
high-level-laws string and the golden content and checking which test
tuples also appear in the training split. This overlap reflects a
structural property of expert legal revisions: distinct violating
provisions, when constrained by the same superior law, are often
resolved by the same or near-identical minimal revision, so the same
$(\superior, \golden)$ tuple recurs across data items by design.

Table~\ref{tab:overlap} summarizes the structure. The hash-based
stratification key is exhaustive over the test split and partitions the
696 test records into 452 Seen-gB (a $(\superior, \golden)$ tuple
identical to at least one training record) and 244 Unseen-gB
(no matching training tuple). The Seen-gB stratum is the one in which
the exact superior--counterfactual pair appeared during Stage~1
pretraining; the Unseen-gB stratum is the structural-disjoint complement
used throughout the stratified audits in
Section~\ref{sec:audits:strat}.\footnote{Table~\ref{tab:overlap}'s
$(\superior, \golden)$ row reports 387 train$\cap$test overlaps as
\emph{distinct overlapping tuples}; the larger 452 Seen-gB count is the
\emph{number of test records affected}, which exceeds 387 because some
training tuples are shared by multiple test records.}

\begin{table}[t]
  \centering
  \small
  \begin{tabular}{lrrr}
    \toprule
    Key & train$\cap$val & train$\cap$test & val$\cap$test \\
    \midrule
    Record id                       &     0 &     0 &     0 \\
    Provision URL                   &   130 &   113 &   108 \\
    Document title                  &   163 &   150 &   148 \\
    $(\superior, \subordinate)$ tuple &    13 &     5 &     0 \\
    $(\superior, \golden)$ tuple    &   745 & \textbf{387} &   219 \\
    \bottomrule
  \end{tabular}
  \caption{Cross-split structure of \dataset{}. Record IDs are
  unique across splits, as expected. URL and title overlap reflects the
  fact that records often quote provisions from the same legal documents.
  Subordinate-provision (super, sub) overlap is small (5 train$\cap$test
  pairs). The $(\superior, \golden)$ tuple overlap is the largest:
  $387$ distinct training $(\superior, \golden)$ tuples are also seen in
  test, affecting $452$ test records (some training tuples recur across
  multiple test records). We use this overlap to define the Seen-gB /
  Unseen-gB stratification reported in
  Section~\ref{sec:audits:strat}.}
  \label{tab:overlap}
\end{table}

The structure also includes moderate URL and title overlap (113 URL,
150 title) across train and test. These reflect a property of the
source corpus: the benchmark draws provisions from the same legal
documents, and many records cite the same statutes. The (super, sub)
overlap of 5 records reflects near-duplicate subordinate provisions;
the (super, $\golden$) overlap of 387 records reflects the convergent
nature of expert legal revisions, in which distinct violating
provisions under the same superior-law constraint are often resolved by
the same minimal revision. The (super, $\golden$) overlap is therefore
a structural property of the legal-revision task, and we use it to
define the Seen-gB / Unseen-gB stratification reported alongside the
primary aggregate result (Section~\ref{sec:audits:strat}).

\subsection{Evaluation protocol}
\label{sec:task-data:eval}

We follow the protocol established by \citet{zhao2026lcrcn}: the model
is trained on the 4828-record training pool with 5-class cross-entropy,
validated on the 1404-record validation pool with macro-F$_1$ for
best-epoch selection, and evaluated once on the 696-record held-out
test pool. Macro-F$_1$ is computed with the locked label set
$\{0,1,2,3,4\}$ and \texttt{zero\_division=0} so that bootstrap resamples
that drop the rare Definition class do not change the denominator.
The pre-registered analysis script
(Section~\ref{sec:prereg:rule}) enforces this label set for all
confirmatory statistics.

Beyond the aggregate macro-F$_1$, we report per-class F$_1$ in the
appendix and the cold-start stratified projection
(Section~\ref{sec:audits:strat}) as a difficulty stratification along
the $(\superior, \golden)$ novelty axis described in
Section~\ref{sec:task-data:audit}. We do not modify the official
test split.


\section{Method: The \method{} Family}
\label{sec:method}

\subsection{Overview}
\label{sec:method:overview}

The \method{} family is a two-stage method for taxonomic legislative
conflict classification. Stage~1 pretrains a shared encoder via a typed
counterfactual selective intervention pretraining objective (CSIP) over
the triplet $(\superior, \subordinate, \golden)$, where $\superior$ is a
superior legal provision, $\subordinate$ is a subordinate provision
under review, and $\golden$ is an expert-written minimal revision of
$\subordinate$ that removes the conflict. Stage~2 transfers the
pretrained encoder to the downstream five-way classification head that
the official \dataset{} benchmark scores.

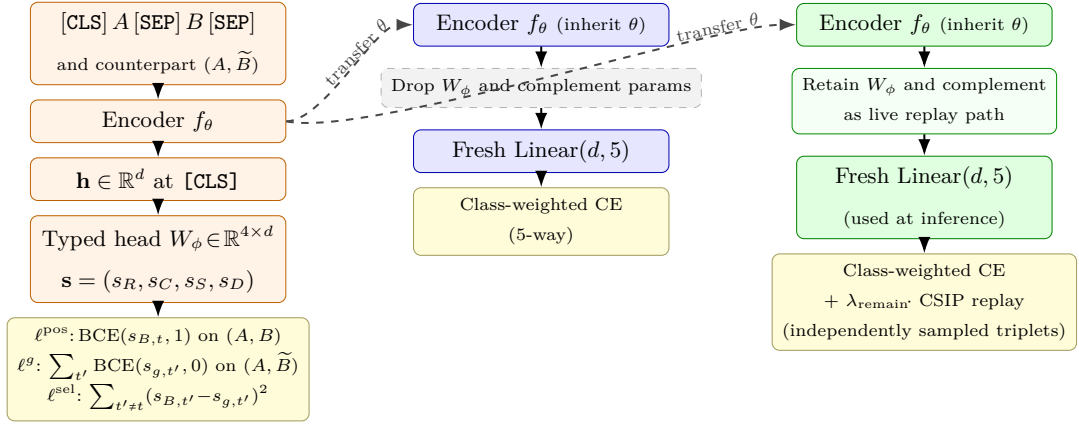
\begin{figure}[t]
  \centering
  \resizebox{\linewidth}{!}{%
  \begin{tikzpicture}[
    font=\small,
    every node/.style={align=center},
    box/.style={draw, rounded corners=3pt, inner sep=4pt,
                minimum width=3.6cm, minimum height=0.6cm},
    s1/.style={box, fill=orange!10, draw=orange!75!black},
    v1/.style={box, fill=blue!10, draw=blue!70!black},
    v2/.style={box, fill=green!12, draw=green!50!black},
    loss/.style={box, fill=yellow!18, draw=yellow!55!black,
                 font=\scriptsize, minimum height=0.95cm},
    drop/.style={box, fill=gray!10, draw=gray!60, font=\scriptsize,
                 dashed, minimum height=0.55cm},
    keep/.style={box, fill=green!4, draw=green!55!black,
                 font=\scriptsize, minimum height=0.55cm},
    title/.style={font=\bfseries\small},
    arr/.style={-Latex, thick},
    transfer/.style={-Latex, thick, dashed, gray!55!black}
  ]
  \node[title, orange!75!black] (TA) at (0,0)    {Stage~1 — Typed CSIP pretraining};
  \node[title, blue!75!black]   (TB) at (5.4,0)  {Stage~2 (\vone): typed-discard};
  \node[title, green!45!black]  (TC) at (10.8,0) {Stage~2 (\vtwo): anti-forget replay};

  \node[s1, below=2mm of TA] (A1)
    {$[\texttt{CLS}]\,\superior\,[\texttt{SEP}]\,\subordinate\,[\texttt{SEP}]$\\
     \scriptsize and counterpart $(\superior,\golden)$};
  \node[s1, below=2mm of A1] (A2) {Encoder $f_\theta$};
  \node[s1, below=2mm of A2] (A3) {$\mathbf{h}\in\mathbb{R}^d$ at \texttt{[CLS]}};
  \node[s1, below=2mm of A3] (A4)
    {Typed head $W_\phi\!\in\!\mathbb{R}^{4\times d}$\\
     $\mathbf{s}=(s_R, s_C, s_S, s_D)$};
  \node[loss, below=2mm of A4] (A5)
    {$\ell^{\text{pos}}{:}\,\text{BCE}(s_{B,t},1)\ \text{on}\ (\superior,\subordinate)$\\
     $\ell^{g}{:}\,\sum_{t'}\text{BCE}(s_{g,t'},0)\ \text{on}\ (\superior,\golden)$\\
     $\ell^{\text{sel}}{:}\,\sum_{t'\neq t}(s_{B,t'}{-}s_{g,t'})^{2}$};
  \draw[arr] (A1) -- (A2);
  \draw[arr] (A2) -- (A3);
  \draw[arr] (A3) -- (A4);
  \draw[arr] (A4) -- (A5);

  \node[v1, below=2mm of TB] (B1) {Encoder $f_\theta$\ \scriptsize(inherit $\theta$)};
  \node[drop, below=3mm of B1] (B2) {Drop $W_\phi$ and complement params};
  \node[v1, below=3mm of B2] (B3) {Fresh $\mathrm{Linear}(d,5)$};
  \node[loss, below=2mm of B3] (B4) {Class-weighted CE\\(5-way)};
  \draw[arr] (B1) -- (B2);
  \draw[arr] (B2) -- (B3);
  \draw[arr] (B3) -- (B4);

  \node[v2, below=2mm of TC] (C1) {Encoder $f_\theta$\ \scriptsize(inherit $\theta$)};
  \node[keep, below=3mm of C1] (C2) {Retain $W_\phi$ and complement\\as live replay path};
  \node[v2, below=3mm of C2] (C3) {Fresh $\mathrm{Linear}(d,5)$\\
                                   \scriptsize(used at inference)};
  \node[loss, below=2mm of C3] (C4)
    {Class-weighted CE \\
     $+\ \lambda_{\text{remain}}\!\cdot$ CSIP replay\\
     (independently sampled triplets)};
  \draw[arr] (C1) -- (C2);
  \draw[arr] (C2) -- (C3);
  \draw[arr] (C3) -- (C4);

  \draw[transfer] (A2.east) to[out=0,in=180,looseness=0.6]
    node[above,sloped,font=\scriptsize,inner sep=1pt,pos=0.6]{transfer $\theta$}
    (B1.west);
  \draw[transfer] (A2.east) to[out=-12,in=180,looseness=0.4]
    node[above,sloped,font=\scriptsize,inner sep=1pt,pos=0.8]{transfer $\theta$}
    (C1.west);
  \end{tikzpicture}%
  }
  \caption{\method{} architecture overview. Stage~1 pretrains the
  encoder with a typed CSIP loss whose head has four rows (one per
  conflict factor) and whose No-Conflict prediction is derived via a
  monotone complement of the per-factor evidence (Eqs.~\ref{eq:l-factor},
  \ref{eq:l-nc}). \vone{}'s Stage~2 discards the typed factor head and
  the monotone-complement parameters and fine-tunes a fresh
  $\mathrm{Linear}(d, 5)$ classifier with class-weighted cross-entropy.
  \vtwo{}'s Stage~2 retains the typed-head parameter group as a live
  replay path on independently sampled CSIP triplets, alongside the
  same fresh $\mathrm{Linear}(d, 5)$ used at inference.}
  \label{fig:method}
\end{figure}

We study two variants of the Stage~2 transfer:
\vone{} (typed-discard), in which the typed factor head and its
monotone-complement parameters from Stage~1 are removed before
fine-tuning, and \vtwo{} (anti-forget replay), in which the typed factor
head is retained alongside a fresh classification head and the CSIP loss
is replayed as a stage-2 auxiliary term on independently sampled
triplets. The two variants share the same Stage~1 procedure; they differ
only in Stage~2. Figure~\ref{fig:method} shows the overall architecture;
Algorithm~\ref{alg:csip} gives the Stage~1 training loop.

\subsection{Pair encoding}
\label{sec:method:encoding}

For an input pair $(\superior, \subordinate)$, the tokenizer constructs
a single BERT sentence-pair sequence
$[\texttt{CLS}]\ \superior\ [\texttt{SEP}]\ \subordinate\ [\texttt{SEP}]$,
truncated longest-first to $512$ tokens. The encoder
$f_\theta(\cdot)$ produces a single contextualized representation
$\mathbf{h} \in \mathbb{R}^d$ at the \texttt{[CLS]} position
($d = 768$). We compute all pair-level quantities below from this
single joint encoding; no separate forward passes on $\superior$ and
$\subordinate$ are taken and no concatenation of two encoder outputs is
constructed. The same encoding scheme is used in all Stage~1 and Stage~2
forward paths and for both encoder backbones.

\subsection{Typed factor head and monotone-complement classifier}
\label{sec:method:head}

For the four typed conflict categories (Responsibility, Condition,
Sanction, Definition) we attach a typed factor head
$W_\phi \in \mathbb{R}^{4 \times d}$ that maps the \texttt{[CLS]}
representation to a raw factor score vector
\begin{equation}
\mathbf{s}(\superior, \subordinate)
    = W_\phi\,\mathrm{Dropout}\!\left(f_\theta(\superior, \subordinate)\right)
    \;\in\;\mathbb{R}^{4} .
\label{eq:s-factor}
\end{equation}
The bias of $W_\phi$ is initialized to $\mathrm{logit}(0.05) \approx
-2.94$ (a low-evidence prior on each factor) and its weight is
initialized with Xavier-uniform, gain $0.5$. We then form a per-factor
evidence vector
$\mathbf{e}(\mathbf{s}) = \sigma(\mathbf{s}) \in [0,1]^4$, where
$\sigma$ is the sigmoid.

To obtain a five-way classification probability that respects the
domain constraint ``No-Conflict is the absence of any factor evidence,''
the typed factor head is paired with a monotone-complement classifier
\begin{align}
\ell_t   &= b_t + \exp(\log w_t)\cdot \mathbf{e}_t,
   \quad t \in \{R, C, S, D\}, \label{eq:l-factor}\\
\ell_{\textrm{NC}} &= b_{\textrm{NC}}
   - \exp(\log \alpha)\cdot \sum_{t=1}^{4} \mathbf{e}_t, \label{eq:l-nc}
\end{align}
producing the five-way logit vector
$\boldsymbol{\ell} = [\ell_R, \ell_C, \ell_S, \ell_D, \ell_{\textrm{NC}}]$.
The per-factor scalars $w_t$ and $\alpha$ are parameterized through
$\exp(\cdot)$ to guarantee positivity; this is necessary so that
$\ell_{\textrm{NC}}$ is a strictly decreasing function of every factor's
evidence, i.e.\ \emph{any} factor firing pushes the No-Conflict (NC)
logit down. We initialize $b_t = 0$, $b_{\textrm{NC}} = +0.5$ (positive
margin so the NC class wins on neutral evidence), $\log\alpha = 0$ (i.e.\ $\alpha =
1$), and $\log w_t = 0$ for all $t$ (i.e.\ $w_t = 1$).

Together,
$\{W_\phi, b_t, \log w_t, b_{\textrm{NC}}, \log\alpha\}$ constitute the
\emph{typed-head parameter group}. The Stage~2 variants differ in
whether this group is preserved or discarded
(Sections~\ref{sec:method:stage2-v1}, \ref{sec:method:stage2-v2}).

\subsection{Stage~1: typed CSIP pretraining}
\label{sec:method:stage1}

Each Stage~1 mini-batch mixes conflicting and no-conflict pairs. For a
conflicting pair with conflict-type label $t \in \{1, 2, 3, 4\}$, the
loader emits both the conflicting pair $(\superior, \subordinate)$ and
its expert-revised counterpart $(\superior, \golden)$, yielding two raw
factor score vectors $\mathbf{s}_B = \mathbf{s}(\superior,
\subordinate)$ and $\mathbf{s}_g = \mathbf{s}(\superior, \golden)$ from
two joint encodings. The per-record CSIP loss has three components:
\begin{align}
\ell^{\textrm{pos}}
   &= \mathrm{BCE}\!\left(\mathbf{s}_{B,t},\, 1\right), \label{eq:l-pos}\\
\ell^{g}
   &= \sum_{t'=1}^{4}
       \mathrm{BCE}\!\left(\mathbf{s}_{g,t'},\, 0\right), \label{eq:l-g}\\
\ell^{\textrm{select}}
   &= \sum_{t' \neq t}
       \bigl(\mathbf{s}_{B,t'} - \mathbf{s}_{g,t'}\bigr)^{2}, \label{eq:l-select}
\end{align}
where $\mathrm{BCE}(x, y)$ is the binary cross-entropy with logits and
target $y$. Equation~\ref{eq:l-pos} requires only the true factor of
$(\superior, \subordinate)$ to fire; Equation~\ref{eq:l-g} forces all
four factors of the expert-revised counterpart to be silent;
Equation~\ref{eq:l-select} penalizes the encoder for changing the
non-target factor scores between $\subordinate$ and $\golden$,
localizing the learned conflict signal to the target factor only.
For a no-conflict pair $(\superior, \subordinate_{\textrm{NC}})$, the
loader emits only the pair itself and the per-record loss is
\begin{equation}
\ell^{\textrm{nc}}
   = \sum_{t'=1}^{4}
       \mathrm{BCE}\!\left(\mathbf{s}_{B_{\textrm{NC}}, t'},\, 0\right). \label{eq:l-nc-bce}
\end{equation}
The Stage-1 batch loss averages per-record losses:
\begin{equation}
\mathcal{L}_{\textrm{CSIP}} =
   \frac{1}{|\mathcal{B}|}
   \sum_{r \in \mathcal{B}}
   \begin{cases}
     \ell^{\textrm{pos}}_r + \ell^{g}_r
       + \lambda_{\textrm{select}}\cdot \ell^{\textrm{select}}_r
       & r\ \textrm{conflict},\\
     \ell^{\textrm{nc}}_r
       & r\ \textrm{NC}.
   \end{cases}
\label{eq:l-csip}
\end{equation}
We fix $\lambda_{\textrm{select}} = 1.0$ throughout, locked in the
pre-registration (Section~\ref{sec:prereg}).

This formulation differs from softmax cross-entropy over the five
classes in three concrete ways: (i) the typed head has four rows, one
per factor, and the No-Conflict class is derived via the monotone
complement in Equation~\ref{eq:l-nc} rather than predicted as a fifth
logit; (ii) the supervision signal on $\golden$ is a zero-target on
\emph{all} four factors, not a softmax target on a fifth No-Conflict
class; (iii) the selectivity term is a squared difference on the
\emph{non}-target factors, not a divergence between two distributions
over conflict types. Compared to the prefix-tuned counterfactual
contrastive objective of \citet{ccprefix2024} and the medical-imaging
counterfactual contrastive learning of \citet{roschewitz2024cfsimclr},
our supervision uses domain-expert revisions $\golden$ as the
counterfactual positives rather than synthesized images or prefix
tokens, and the inductive bias of the monotone-complement classifier
encodes a hand-specified factor-additive structure of legislative
conflict.

\subsection{Stage~2 (\vone{}): typed-discard transfer}
\label{sec:method:stage2-v1}

After Stage~1 converges, the \vone{} transfer \emph{discards} the entire
typed-head parameter group $\{W_\phi, b_t, \log w_t,
b_{\textrm{NC}}, \log\alpha\}$ along with the monotone-complement
classifier. A fresh five-way head $W_{\textrm{cls}} \in
\mathbb{R}^{5 \times d}$ is initialized with the default
\texttt{nn.Linear} initialization (Kaiming-uniform). The Stage~2 forward
path on a sentence pair $(\superior, \subordinate)$ is
\begin{equation}
\hat{\mathbf{y}}(\superior, \subordinate) =
   W_{\textrm{cls}}\,\mathrm{Dropout}\!\left(
      f_\theta(\superior, \subordinate)
   \right),
\label{eq:v1-forward}
\end{equation}
and the loss is class-weighted cross-entropy with inverse-frequency
weights on the five-way taxonomic label of the downstream task. No CSIP
loss participates in \vone{} Stage~2; the only signal passed from
Stage~1 is the encoder weights.

The typed-discard transfer is the same projection-head-discard pattern
used in contrastive vision pretraining \citep{chen2020simclr}; the
difference is that our discarded structure is the typed factor head
plus its monotone-complement classifier, not a generic projection MLP.

\subsection{Stage~2 (\vtwo{}): anti-forget replay}
\label{sec:method:stage2-v2}

The \vtwo{} variant is motivated by the continual-learning hypothesis
that vanilla CE fine-tuning erases the conflict-type discrimination
structure that Stage~1 installed in the encoder. Instead of discarding the typed-head
parameter group, \vtwo{} retains every parameter of the group as a live
group in the fine-tuning optimizer (same Python objects, same
\texttt{nn.Parameter} instances, gradient flow into the Stage-1 weights
preserved through the replay term), and adds the same fresh head
$W_{\textrm{cls}}$ used in \vone{} as the only inference-time
classifier.

Each fine-tuning step now consumes two interleaved mini-batches: a
fine-tuning (FT) batch of $(\superior, \subordinate, y)$ pairs with five-way labels and
a CSIP batch of $(\superior, \subordinate, \golden, t)$ triplets drawn
from an independently shuffled triplet loader. The FT batch flows
through $W_{\textrm{cls}}$ (Equation~\ref{eq:v1-forward}) and yields the
cross-entropy term $\mathcal{L}_{\textrm{CE}}$; the CSIP batch flows
through the retained typed head (Equation~\ref{eq:s-factor}) and yields
the replay term $\mathcal{L}_{\textrm{CSIP}}^{\textrm{replay}}$, defined
identically to Equation~\ref{eq:l-csip} but evaluated on the
fine-tuning-stage CSIP loader. The Stage~2 loss is
\begin{equation}
\mathcal{L}_{\vtwo{}} =
   \mathcal{L}_{\textrm{CE}}
   + \lambda_{\textrm{remain}} \cdot
       \mathcal{L}_{\textrm{CSIP}}^{\textrm{replay}},
\label{eq:l-v2}
\end{equation}
with $\lambda_{\textrm{remain}} = 0.5$ fixed a priori in the
pre-registration. The fresh head and the retained typed head receive
gradient from their respective losses; the encoder receives gradient
from both via the shared representation. Inference uses only
$W_{\textrm{cls}}$; the typed-head parameters remain in
\texttt{state\_dict()} but do not contribute to the prediction.

\paragraph{Transparency note on architectural retention.}
The Stage-2 architecture of \vtwo{} differs from \vone{} not only in the
inclusion of the replay loss but also in the train-time module class:
the full typed-head parameter group is preserved as a live group rather
than discarded. We interpret the confirmatory result for \vtwo{} as
evidence for the pre-registered redesign as a whole (the replay loss
plus the train-time head configuration required to implement it), not
for a loss-only modification in isolation. Our matched-seed audit
(Section~\ref{sec:audits:v1}) directly compares \vone{} against
\vtwo{} on the same 18 pre-registered seeds to bound the marginal
contribution of this redesign; a factorial ablation that varies the
loss and the head retention independently is left to follow-up work
(Section~\ref{sec:limitations}).

\subsection{Training procedure}
\label{sec:method:training}

We pretrain Stage~1 with the AdamW optimizer at learning rate
$2{\times}10^{-5}$, weight decay $0.01$, linear warmup with
$\textrm{warmup\_ratio} = 0.1$ and a linear decay schedule, gradient
clipping at norm $1.0$, batch size $32$ on the CSIP triplet loader,
for $3$ epochs. The Stage-2 fine-tuning runs for $5$ epochs at the same
base learning rate with FT batch size $16$. For \vtwo{}, the secondary
CSIP-replay loader uses batch size $8$, set to keep the two concurrent
encoder forward passes (FT and replay) within the $24$~GB VRAM budget
of a single RTX~3090; this stage-2 replay batch size was fixed a priori
in the pre-registration after a pre-launch out-of-memory (OOM) remediation
(Section~\ref{sec:prereg}). Class-weighted cross-entropy with
inverse-frequency weights addresses the Definition class imbalance
($n = 48$ in test; the rarest of the five). The validation split selects
the best epoch by macro-F$_1$; the test split is held out for the
single locked confirmatory evaluation. All hyperparameters are
summarized in Table~\ref{tab:hyperparameters}.

\begin{table}[t]
\centering
\small
\begin{tabular}{lll}
\toprule
Hyperparameter & Stage 1 (CSIP) & Stage 2 (FT) \\
\midrule
Optimizer            & AdamW & AdamW \\
Learning rate        & $2{\times}10^{-5}$ & $2{\times}10^{-5}$ \\
Weight decay         & $0.01$ & $0.01$ \\
Warmup ratio         & $0.1$ (linear) & $0.1$ (linear) \\
LR schedule          & linear decay & linear decay \\
Gradient clip (norm) & $1.0$ & $1.0$ \\
Dropout              & $0.1$ (encoder + heads) & $0.1$ \\
Max sequence length  & $512$ (joint pair) & $512$ (joint pair) \\
Batch size           & $32$ (triplet) & $16$ (FT) \\
Replay batch         & --- & $8$ (\vtwo{} only) \\
Epochs               & $3$ & $5$ \\
Loss coefficient     & $\lambda_{\textrm{select}} = 1.0$ & $\lambda_{\textrm{remain}} = 0.5$ (\vtwo{}) \\
Class weights        & --- & inverse frequency \\
Model-selection metric & --- & val macro-F$_1$ \\
\bottomrule
\end{tabular}
\caption{\method{} hyperparameters, all fixed a priori in the v6
pre-registration (Section~\ref{sec:prereg}). The replay batch size
of $8$ for \vtwo{} stage-2 was set after a pre-launch OOM remediation
documented in the pre-reg amendment; all other values are unchanged
from v1. Both backbones (\texttt{chinese-roberta-wwm-ext},
\texttt{SAILER}) use the same hyperparameter table.}
\label{tab:hyperparameters}
\end{table}

We instantiate the encoder $f_\theta$ from two backbones:
\texttt{chinese-roberta-wwm-ext} \citep{cui2020chinesebert} as the
primary backbone, and SAILER \citep{li2023sailer} as a cross-backbone
replication. Both are 12-layer BERT-architecture encoders. SAILER is
pretrained on Chinese legal corpora and provides the cross-domain check
of whether our gains depend on the general-Chinese vs.\ legal-Chinese
pretraining of the backbone.

\begin{algorithm}[t]
\caption{Stage~1 typed CSIP pretraining (single epoch).}
\label{alg:csip}
\begin{algorithmic}[1]
\State \textbf{Input:} encoder $f_\theta$, typed-head group
   $\{W_\phi, b_t, \log w_t, b_{\textrm{NC}}, \log\alpha\}$, conflict
   pairs $\mathcal{D}_c = \{(\superior_i, \subordinate_i, \golden_i,
   t_i)\}$, no-conflict pairs $\mathcal{D}_{nc} =
   \{(\superior_j, \subordinate_{\textrm{NC},j})\}$
\State \textbf{Output:} updated encoder + typed-head parameters
\For{each mini-batch $\mathcal{B} \sim \mathcal{D}_c \cup \mathcal{D}_{nc}$}
   \State $\mathcal{L} \gets 0$
   \For{each conflicting record $(\superior_i, \subordinate_i, \golden_i, t_i) \in \mathcal{B}$}
     \State $\mathbf{h}_B \gets f_\theta(\superior_i, \subordinate_i)$
        \Comment{single joint encoding}
     \State $\mathbf{h}_g \gets f_\theta(\superior_i, \golden_i)$
     \State $\mathbf{s}_B \gets W_\phi\,\mathrm{Dropout}(\mathbf{h}_B)$;
            $\mathbf{s}_g \gets W_\phi\,\mathrm{Dropout}(\mathbf{h}_g)$
     \State $\ell^{\textrm{pos}} \gets
            \mathrm{BCE}(\mathbf{s}_{B,t_i},\, 1)$
     \State $\ell^{g}        \gets
            \sum_{t'=1}^{4}\mathrm{BCE}(\mathbf{s}_{g,t'},\, 0)$
     \State $\ell^{\textrm{select}} \gets
            \sum_{t' \neq t_i}(\mathbf{s}_{B,t'} - \mathbf{s}_{g,t'})^{2}$
     \State $\mathcal{L} \gets \mathcal{L} + \ell^{\textrm{pos}}
            + \ell^{g} + \lambda_{\textrm{select}}\cdot \ell^{\textrm{select}}$
   \EndFor
   \For{each no-conflict record $(\superior_j, \subordinate_{\textrm{NC},j}) \in \mathcal{B}$}
     \State $\mathbf{h} \gets f_\theta(\superior_j, \subordinate_{\textrm{NC},j})$
     \State $\mathbf{s} \gets W_\phi\,\mathrm{Dropout}(\mathbf{h})$
     \State $\mathcal{L} \gets \mathcal{L}
            + \sum_{t'=1}^{4}\mathrm{BCE}(\mathbf{s}_{t'},\, 0)$
   \EndFor
   \State $\mathcal{L} \gets \mathcal{L} / |\mathcal{B}|$
   \State Backpropagate $\mathcal{L}$; AdamW step on encoder + typed-head
\EndFor
\end{algorithmic}
\end{algorithm}

\section{Pre-registration Protocol}
\label{sec:prereg}

\subsection{Cryptographic commitment and seed derivation}
\label{sec:prereg:osf}

The confirmatory study reported in Section~\ref{sec:experiments} was
registered on the Open Science Framework (OSF) under project~3ye4c. The v6
pre-registration file, locked on 2026-05-15 before any v6 measurement
existed, has SHA-256 fingerprint
\texttt{f1124624\ldots}.\footnote{Full SHA omitted for anonymization; the
verbatim hash appears in the registered file.} The pre-registration commits
the cells, backbones, seeds, training scripts, analysis script, decision
rule, and missingness policy in a single immutable record. OSF assigns a
trusted timestamp; revisions create new records rather than overwriting the
original, so the time-ordering between the registered design and the
observed data is verifiable by any reviewer.

The eighteen primary seeds and twelve backup seeds were not chosen by the
authors. They were derived by a deterministic PRNG snippet seeded from the
first eight hex digits of an earlier, independently OSF-anchored
pre-registration's SHA hash (specifically, the v5 pre-reg SHA on
osf.io/e57xn): \texttt{int("8607bca5", 16) = 2248653989}. The snippet draws
disjoint sets from a candidate pool that excludes all previously used seeds
(v1 + v5), then verifies that no selected seed appears in a banned-seed list.
The whole derivation appears verbatim in the v6 pre-reg file. A reviewer
can re-execute the snippet against the OSF-hosted v5 file to reproduce the
seed lists byte-identically. This rules out post-hoc seed selection as a
source of inflation.

\subsection{Locked decision rule}
\label{sec:prereg:rule}

The confirmatory test compares each method cell to the standard concatenated
classification baseline (C2) on the official $N{=}696$ test split. The
primary estimator is seed-as-unit, which resamples the eighteen pre-reg
seeds rather than test examples \citep{efron1979bootstrap, koehn2004bootstrap}.
This estimator captures the training-stochasticity uncertainty that the
claim "method beats baseline in expectation over training" requires.

The locked decision rule C1 requires \emph{all three} of the following:
(a) the mean per-seed difference in macro-F$_1$ is at least 0.8 percentage
points; (b) the lower bound of the 95\% seed-bootstrap CI (20{,}000 rounds,
RNG~seed~4242) exceeds zero; (c) the lower bound of the Student-$t$ 95\%
interval (df~$={}$17) exceeds zero. The both-intervals requirement
yields a single decision under two interval estimators. A
secondary descriptive label C1' (mean $\geq$ 1.0~pp and both lower bounds
positive) is reported for completeness but never upgrades a primary FAIL.
The same locked rule and analysis script apply to both backbones.

Both the analysis script (SHA \texttt{7d9d67a3\ldots}) and its invocation
arguments are committed in the pre-reg. All \texttt{f1\_score} calls in the
confirmatory analysis pass \texttt{labels=[0,1,2,3,4],
zero\_division=0}, so a bootstrap resample that drops the rare Definition
class cannot change the macro-F$_1$ denominator. Strict ID/size invariants
on the test split (exactly 696 unique IDs, identical gold-label vector
across cells and seeds) are enforced at analysis time; any deviation
aborts the run.

\subsection{Fixed-$N$, no-extension commitment}
\label{sec:prereg:fixed-n}

The pre-registration fixes $N{=}18$ seeds per (backbone $\times$ cell) and
forbids extension. If the primary RoBERTa result fails the locked rule, the
project commits to closing the entire \method{} line as a primary
contribution candidate (the family-closure commitment, registered as
\S0a). No additional seeds may be run to convert a marginal failure into a
pass after the fact \citep{steegen2016multiverse, simmons2011falsepositive}.

\subsection{Staged execution and missingness policy}
\label{sec:prereg:staged}

To avoid burning compute on a secondary backbone when the primary fails,
the pre-reg \S0c specifies staged execution: Stage~A runs the primary
backbone (RoBERTa) over all eighteen seeds; the locked rule is evaluated
once on Stage~A; Stage~B (SAILER) is invoked iff Stage~A passes. The
Stage~A--Stage~B gate consults only the locked decision rule, no
intermediate diagnostics. A single Python orchestrator enforces the
restart/backup-substitution state machine without operator discretion: an
infrastructure failure (matching a metric-free regular expression for
out-of-memory, CUDA error, host disconnect) retries the same seed up to
twice; a training failure substitutes the next unused backup seed and
reruns from scratch with the new seed. The backup substitution preserves
$N{=}18$ valid runs per cell.

\subsection{Provenance and integrity}
\label{sec:prereg:provenance}

A no-peek discipline prevents outcome-conditioned changes during execution:
the operator monitors process state and disk only, with metric strings
redacted by a regular-expression filter. The pre-reg, the four script
SHA-256 hashes, and a post-results addendum describing two auxiliary
analyses added after the primary result was observed (the matched-seed
\vone{} rerun in Section~\ref{sec:audits:v1} and the cold-start
stratified evaluation in Section~\ref{sec:audits:strat}) are all archived
on OSF. An integrity checklist covering citations, numbers, methodology,
and narrative framing was run twice against the manuscript, once after
the full draft and once again in a fresh verification context.

\paragraph{Transparency note on framing evolution.}
The OSF timestamp record shows a v5 pre-registration (2026-05-14) preceding
the v6 (2026-05-15). The v5 confirmatory test of \vone{} marginally failed
the same locked rule on the primary RoBERTa backbone at the smaller fixed
size $N{=}9$. The v6 redesign committed in advance to two changes: a
doubled fixed size ($N{=}9 \to 18$) and the \vtwo{} anti-forget
replay\footnote{The \vtwo{} redesign retains the typed factor head as a
train-time live parameter group, in addition to adding the CSIP replay
term; this architectural retention is part of the locked v6 design (see
Section~\ref{sec:method:stage2-v2}).}. The matched-seed audit in
Section~\ref{sec:audits:v1} was added \emph{after} the v6 confirmatory
result, explicitly to isolate the contribution of the redesign from the
mean lift attributable to seed-set drift, and is reported as auxiliary
rather than confirmatory. The CMCL pre-registration precedent of
\citet{qiu2024cmcl} demonstrates that OSF anchoring of NLP designs
is feasible; we apply the same discipline here without claiming primacy.


\section{Experiments}
\label{sec:experiments}

\subsection{Setup}
\label{sec:experiments:setup}

The confirmatory test compares \vtwo{} (the pre-registered cell) against
C2, our reimplementation of a single-model dense classifier that
concatenates the \texttt{[CLS]} representations of the superior and
subordinate provisions and applies a class-weighted cross-entropy
classifier. C2 matches the dense-encoder baseline family reported in
the dataset paper~\citep{zhao2026lcrcn} and serves as the no-$\golden$
reference point against which the typed CSIP family is compared.

The data splits are the official train / $1404$ validation / $696$ test
partition from the \dataset{} benchmark, reported by
\citet{zhao2026lcrcn} as $4895$ training records pre-filter and
yielding $4828$ training records after our ingest filters
(Section~\ref{sec:task-data:splits}); validation and test splits load
cleanly at their nominal sizes. We use only the 5-class taxonomic
conflict labels (Responsibility, Condition, Sanction, Definition,
No-Conflict). The validation split selects the best-epoch checkpoint
by macro-F$_1$; the test split is held out for the single locked
evaluation. All hyperparameters are listed in
Table~\ref{tab:hyperparameters}.

The primary backbone is \texttt{chinese-roberta-wwm-ext}
\citep{cui2020chinesebert}; the secondary backbone for cross-backbone
replication is SAILER \citep{li2023sailer}. Pre-registration \S0c
specifies staged execution: Stage~A runs RoBERTa across all eighteen
locked seeds; the decision rule is evaluated once; Stage~B (SAILER) is
invoked only if Stage~A passes (Section~\ref{sec:prereg}). All
implementation, data preprocessing, and statistical-analysis scripts are
SHA-anchored in the registered file.

We report the per-seed difference $\Delta_s = \text{mF1}_s(\vtwo{}) -
\text{mF1}_s(C2)$ in percentage points. The primary estimator is
seed-as-unit with two CI variants computed in parallel: a seed-bootstrap
95\% interval (B${=}$20{,}000 rounds, fixed RNG seed~4242) and a
Student-$t$ 95\% interval with $df{=}17$. The locked decision rule C1
requires the mean to clear 0.8~pp \emph{and} both lower bounds to clear
zero; the descriptive label C1$'$ requires mean${\geq}$1.0~pp on top of
C1's both-intervals criterion.

\subsection{Stage~A: Primary backbone (RoBERTa)}
\label{sec:experiments:stage-a}

Stage~A completed on a single RTX~3090 in 11.8~h wall-clock time;
all 18 primary seeds executed without infrastructure failure or backup
substitution. The eighteen per-seed deltas range from $-0.44$ to $+3.18$~pp,
with two seeds yielding small negatives ($6607 = -0.15$~pp, $8176 =
-0.44$~pp). Table~\ref{tab:per-seed} reports the full per-seed values.

Aggregating across the eighteen seeds, \vtwo{} on RoBERTa achieves a mean
difference of $+0.916$~pp in macro-F$_1$ over the C2 baseline. The
seed-bootstrap 95\% interval is $[+0.512,\,+1.346]$~pp and the Student-$t$
95\% interval is $[+0.448,\,+1.384]$~pp. Both lower bounds clear zero and
the mean exceeds the 0.8~pp threshold; the locked decision rule C1
passes. The descriptive C1$'$ label (which requires mean ${\geq}1.0$~pp)
does not apply: 0.916 falls 0.084~pp short of the strength bar. The
secondary, example-conditional bootstrap (which resamples test examples
with seeds fixed) yields a wider interval $[-0.413,\,+2.250]$~pp; per the
pre-registered protocol this secondary cannot upgrade a primary FAIL and
is reported for completeness.

Stage~A's PASS verdict triggered Stage~B per the pre-registered staged
execution gate.

\subsection{Stage~B: Cross-backbone replication (SAILER)}
\label{sec:experiments:stage-b}

Stage~B used the same eighteen seeds and the same training-script SHA, with
the backbone path swapped to SAILER. Stage~B completed in 11.3~h,
with no infrastructure failure or substitution. Per-seed deltas span
$-1.31$ to $+4.60$~pp, with two negative seeds ($3943 = -1.31$~pp,
$7565 = -0.31$~pp). Sixteen of the eighteen SAILER seeds yielded a
positive delta.

The aggregate mean over the eighteen seeds is $+1.288$~pp. The
seed-bootstrap 95\% interval is $[+0.676,\,+1.922]$~pp; the Student-$t$
95\% interval is $[+0.602,\,+1.974]$~pp. The locked decision rule C1
passes (mean ${\geq}0.8$, both lower bounds clear zero). The descriptive
C1$'$ label is met on SAILER (mean exceeds 1.0~pp; both lower bounds
positive). The secondary example-conditional CI is
$[+0.046,\,+2.606]$~pp; again reported only for completeness. The cross-backbone
replication satisfies the pre-registered Scenario~1 of \S14: both
backbones cleared the locked decision rule under fresh PRNG-derived seeds
with no post-hoc adjustments.

The cross-backbone result also shows descriptive heterogeneity.
SAILER's per-seed standard deviation ($1.380$~pp) exceeds RoBERTa's
($0.941$~pp), and the SAILER delta has both higher upside and lower
worst-case than RoBERTa. We do
not attempt to attribute this to the legal-pretrained character of the
SAILER backbone; the matched-seed audit in Section~\ref{sec:audits:v1}
shows that mean-delta differences across backbones can be confounded with
seed-set and Stage-2 design choices.

\subsection{Per-seed detail}
\label{sec:experiments:per-seed}

Table~\ref{tab:per-seed} presents the full per-seed macro-F$_1$ for the
C2 baseline and the three method cells discussed in this paper:
\vtwo{}-RoBERTa, \vtwo{}-SAILER, and \vone{}-RoBERTa (the matched-seed
auxiliary of Section~\ref{sec:audits:v1}). All four cells share the same
eighteen seeds and the same data preprocessing pipeline. The per-seed
values are the raw best-validation-epoch macro-F$_1$ scores on the held-out
test split; no smoothing, ensembling, or seed averaging precedes the
per-seed entries.

The eighteen seeds are deterministic and immutable: they were derived
before any v6 measurement existed from the PRNG snippet anchored in
the v5 pre-reg SHA (Section~\ref{sec:prereg:osf}). The same seed set
appears across the three method cells, enabling the matched-seed
comparison in Section~\ref{sec:audits:v1}.

\begin{table}[t]
  \centering
  \small
  \resizebox{\textwidth}{!}{%
  \begin{tabular}{rrrrrrrrr}
    \toprule
     & \multicolumn{2}{c}{\vtwo{}-RoBERTa} & \multicolumn{2}{c}{\vtwo{}-SAILER} & \multicolumn{2}{c}{\vone{}-RoBERTa} & \multicolumn{2}{c}{C2-RoBERTa, C2-SAILER} \\
    \cmidrule(lr){2-3}\cmidrule(lr){4-5}\cmidrule(lr){6-7}\cmidrule(lr){8-9}
    Seed & mF1 & $\Delta$ & mF1 & $\Delta$ & mF1 & $\Delta$ & mF1 & mF1 \\
    \midrule
    838  & 84.77 & +2.04 & 84.56 & +2.50 & 84.50 & +1.76 & 82.73 & 82.06 \\
    1189 & 83.31 & +0.13 & 84.52 & +1.63 & 84.17 & +0.99 & 83.18 & 82.89 \\
    1277 & 86.98 & +3.18 & 84.87 & +2.19 & 85.41 & +1.61 & 83.80 & 82.68 \\
    1339 & 83.17 & +0.38 & 85.08 & +0.49 & 84.91 & +2.11 & 82.79 & 84.60 \\
    1584 & 83.32 & +0.47 & 85.85 & +1.91 & 84.61 & +1.75 & 82.85 & 83.95 \\
    1727 & 84.75 & +0.22 & 85.46 & +2.73 & 86.41 & +1.88 & 84.53 & 82.73 \\
    2502 & 83.11 & +0.48 & 83.08 & +0.20 & 85.67 & +3.04 & 82.63 & 82.88 \\
    3943 & 84.77 & +1.40 & 83.93 & $-$1.31 & 83.79 & +0.42 & 83.37 & 85.25 \\
    4202 & 84.89 & +0.95 & 85.11 & +0.54 & 84.03 & +0.09 & 83.94 & 84.56 \\
    4962 & 84.53 & +1.49 & 85.38 & +1.69 & 84.74 & +1.71 & 83.04 & 83.70 \\
    6607 & 84.67 & $-$0.15 & 86.71 & +4.60 & 84.09 & $-$0.73 & 84.82 & 82.11 \\
    7146 & 84.45 & +0.14 & 85.23 & +2.82 & 84.88 & +0.57 & 84.31 & 82.41 \\
    7516 & 84.36 & +0.11 & 84.48 & +0.81 & 85.41 & +1.16 & 84.25 & 83.66 \\
    7565 & 84.89 & +1.73 & 83.15 & $-$0.31 & 84.53 & +1.37 & 83.16 & 83.46 \\
    8176 & 83.44 & $-$0.44 & 84.02 & +0.35 & 84.70 & +0.82 & 83.88 & 83.67 \\
    8635 & 84.29 & +0.99 & 83.34 & +1.25 & 84.61 & +1.31 & 83.30 & 82.09 \\
    9329 & 85.46 & +2.02 & 83.47 & +0.84 & 83.42 & $-$0.02 & 83.43 & 82.63 \\
    9900 & 84.98 & +1.34 & 83.76 & +0.26 & 84.27 & +0.63 & 83.64 & 83.50 \\
    \midrule
    Mean & 84.45 & $+0.916$ & 84.55 & $+1.288$ & 84.67 & $+1.137$ & 83.54 & 83.30 \\
    SD   &       & 0.941   &       & 1.380   &       & 0.897   &       &       \\
    \bottomrule
  \end{tabular}%
  }
  \caption{Per-seed test macro-F$_1$ for the three method cells. Cells
  share the same eighteen PRNG-derived primary seeds. The two C2 columns
  on the right are the matched baselines for each backbone (left for
  RoBERTa, right for SAILER). The bottom row reports the seed-aggregate
  mean per-seed difference (in percentage points) and its standard
  deviation. All cells were trained and evaluated under the locked
  Stage-1/Stage-2 protocol described in Section~\ref{sec:method}.}
  \label{tab:per-seed}
\end{table}

\subsection{Confirmatory result summary}
\label{sec:experiments:summary}

Table~\ref{tab:main} summarizes the four cells discussed in this paper
under the locked decision rule. The pre-registered Stage~A and Stage~B
confirmatory tests both pass the C1 rule. Section~\ref{sec:audits:v1}
reports a matched-seed auxiliary that adds a fourth row, and
Section~\ref{sec:audits:strat} reports a cold-start stratified projection
that complements the table.

\begin{table}[t]
  \centering
  \small
  \resizebox{\textwidth}{!}{%
  \begin{tabular}{lrrrcc}
    \toprule
    Cell                          & Mean $\Delta$ (pp) & Seed-bootstrap 95\% CI & Student-$t$ 95\% CI & C1 & C1$'$ \\
    \midrule
    \vtwo{}-RoBERTa (Stage~A, pre-reg primary) & $+0.916$ & $[+0.512,\,+1.346]$ & $[+0.448,\,+1.384]$ & PASS & --- \\
    \vtwo{}-SAILER (Stage~B, cross-backbone)   & $+1.288$ & $[+0.676,\,+1.922]$ & $[+0.602,\,+1.974]$ & PASS & PASS \\
    \vone{}-RoBERTa (matched-seed auxiliary; \S\ref{sec:audits:v1})   & $+1.137$ & $[+0.732,\,+1.535]$ & $[+0.691,\,+1.583]$ & PASS & PASS \\
    \midrule
    \vone{}-RoBERTa (historical $N{=}9$ pre-reg v5; FAIL, transparency) & $+1.088$ & $[+0.122,\,+2.051]$ & $[-0.098,\,+2.274]$ & FAIL & --- \\
    \bottomrule
  \end{tabular}%
  }
  \caption{Pre-registered confirmatory result summary plus matched-seed
  auxiliary. The C1 rule requires mean${\geq}0.8$~pp and both 95\% CI
  lower bounds ${>}0$; the C1$'$ descriptive label additionally requires
  mean${\geq}1.0$~pp. The historical row is reported for transparency
  (Section~\ref{sec:prereg}); its numbers are drawn from the v5
  pre-registration archived on OSF (registration \texttt{e57xn}), whose
  prediction files are deposited at that record and not re-distributed
  with the v6 supplementary. The v5 fixed-$N{=}9$ test marginally
  failed C1 because the Student-$t$ lower bound missed zero by
  $0.098$~pp. The v6 fixed-$N{=}18$ tests with the locked rule pass on
  both backbones.}
  \label{tab:main}
\end{table}

The pre-registered \vtwo{} cell clears C1 on both backbones with
PRNG-derived seeds unobserved at OSF lock time and with no post-hoc
adjustment to the analysis script, the seeds, or the rule.
Section~\ref{sec:audits:v1} reports a matched-seed auxiliary on the
simpler \vone{} variant.


\section{Stratified and Matched-Seed Audits}
\label{sec:audits}

The confirmatory result in Section~\ref{sec:experiments} establishes that
\vtwo{} clears the locked decision rule on both backbones. Two questions
remain open. First, does the \vtwo{} gain survive on the harder subset
of the test split whose $(\superior, \golden)$ tuple is novel relative
to training, isolating the part of the gain that does not rely on the
recurring structural overlap described in
Section~\ref{sec:task-data:audit}? Second, does the v6 \vtwo{} redesign
actually contribute the observed gain, or is the v6 mean a re-expression
of a gain already attainable with the simpler \vone{} variant on matched
seeds? The two auxiliaries in this section answer these questions. Both were added \emph{after} the v6 confirmatory
result was observed and are reported as auxiliary, not as new
confirmatory tests; the locked decision rule of
Section~\ref{sec:prereg:rule} continues to apply only to the v6
confirmatory data.

\subsection{Cold-start stratified evaluation}
\label{sec:audits:strat}

We partition the 696-example test split by overlap of the
$(\superior, \golden)$ key with the training split. A test item lies in
\emph{Unseen-gB} if no training item shares both its superior provision
and its expert revision $\golden$; otherwise it lies in \emph{Seen-gB}.
The partition uses MD5 hashes of the joined high-level laws (the superior
identifier) and the golden subordinate text, and is byte-identical to the
hash function we use to identify duplicate revisions. The partition
yields $n{=}244$ Unseen-gB and $n{=}452$ Seen-gB. This stratification
removes exact $(\superior, \golden)$-tuple reuse but does not remove
$\superior$-only overlap; we discuss the residual confound in
Section~\ref{sec:limitations}.

For each backbone we project the per-seed predictions of \vtwo{} and C2
onto the two strata and recompute the seed-as-unit mean and the two
95\% CIs over the eighteen pre-registered seeds. Macro-F$_1$ uses the
same locked label set as the primary analysis. Table~\ref{tab:strat}
reports the result.

\begin{table}[t]
  \centering
  \small
  \begin{tabular}{llrrr}
    \toprule
    Backbone & Stratum ($n$) & Mean $\Delta$ (pp) & Seed-bootstrap 95\% CI & Student-$t$ 95\% CI \\
    \midrule
    RoBERTa  & Full test (696) & $+0.916$ & $[+0.512,\,+1.346]$ & $[+0.448,\,+1.384]$ \\
             & Unseen-gB (244) & $+3.136$ & $[+1.834,\,+4.318]$ & $[+1.754,\,+4.519]$ \\
             & Seen-gB (452)   & $-0.305$ & $[-0.797,\,+0.201]$ & $[-0.862,\,+0.251]$ \\
    \midrule
    SAILER   & Full test (696) & $+1.288$ & $[+0.676,\,+1.922]$ & $[+0.602,\,+1.974]$ \\
             & Unseen-gB (244) & $+2.350$ & $[+1.113,\,+3.510]$ & $[+1.016,\,+3.683]$ \\
             & Seen-gB (452)   & $+0.820$ & $[+0.118,\,+1.599]$ & $[-0.004,\,+1.645]$ \\
    \bottomrule
  \end{tabular}
  \caption{Cold-start stratified projection of the \vtwo{}-vs-C2 per-seed
  predictions onto Unseen-gB and Seen-gB. The Unseen-gB stratum has both
  CI lower bounds above zero on both backbones. The Seen-gB stratum shows
  a negative point estimate on RoBERTa and an interval that crosses zero
  on SAILER. The CI computation matches the locked analysis script byte
  for byte except for the projection onto the stratum.}
  \label{tab:strat}
\end{table}

Under the conservative cold-start stratification that removes all test
items whose exact $(\superior, \golden)$ tuple appears in training, the
\vtwo{} gain remains positive with both confidence intervals above zero
on both backbones. The complementary Seen-gB slice does not show a larger
effect, which is inconsistent with an exact-tuple shortcut account of
the full-test gain.

We do not interpret the Unseen-gB-vs-Seen-gB inequality mechanistically.
The two strata have meaningfully different class composition: the
Unseen-gB stratum has a higher fraction of No-Conflict examples
(reflected in the higher absolute macro-F$_1$ on both backbones), while
Seen-gB concentrates the rare conflict classes (lower absolute
macro-F$_1$). The within-stratum delta is informative about
within-stratum reproducibility, not about which stratum is "harder" in
an absolute sense. We report the cross-stratum heterogeneity
descriptively and treat it as evidence against an exact-tuple shortcut
account of the gain, not as a causal claim about which stratum is
intrinsically harder.

\paragraph{Per-class breakdown of the Unseen-gB gain.}
Table~\ref{tab:strat-perclass} decomposes the Unseen-gB delta into
per-class macro-F$_1$ differences. On both backbones the Unseen-gB
gain is concentrated in the conflict-type classes: Definition
($+8.69$~pp on RoBERTa, $+5.18$~pp on SAILER, both with $t$ 95\%
lower bound above zero), Condition ($+3.15$~pp on both backbones,
both with $t$ 95\% lower bound above zero), and Sanction ($+2.74$~pp
on RoBERTa, $+2.62$~pp on SAILER). The No-Conflict class shows a mean
delta of $-0.02$~pp on RoBERTa and $0.00$~pp on SAILER. Because the
Unseen-gB stratum is composed of $50\%$ No-Conflict items by count,
a reading on which the $+3.1$~pp aggregate Unseen-gB gain is an
artifact of the Unseen-gB stratum's NC-heavy composition would
predict a positive NC-class delta as the dominant contributor; the
observed near-zero NC delta is the opposite. Conflict classes drive
the Unseen-gB gain.

\begin{table}[t]
  \centering
  \small
  \begin{tabular}{lccc}
    \toprule
    Class (Unseen-gB count) & RoBERTa $\Delta$ (pp) & SAILER $\Delta$ (pp) \\
    \midrule
    Responsibility ($n{=}37$)  & $+1.12$           & $+0.79$            \\
    Condition      ($n{=}43$)  & $+3.15^{\dagger}$ & $+3.16^{\dagger}$  \\
    Sanction       ($n{=}17$)  & $+2.74$           & $+2.62$            \\
    Definition     ($n{=}25$)  & $+8.69^{\dagger}$ & $+5.18^{\dagger}$  \\
    No-Conflict    ($n{=}122$) & $-0.02$           & $\phantom{+}0.00$ \\
    \bottomrule
  \end{tabular}
  \caption{Per-class mean macro-F$_1$ delta within the Unseen-gB
  stratum ($\vtwo{}$~minus~C2, averaged over the 18 pre-registered
  seeds). $^{\dagger}$~Student-$t$ 95\% CI lower bound above zero.
  Both backbones concentrate the Unseen-gB gain on Definition,
  Condition, and Sanction; No-Conflict is essentially unchanged.}
  \label{tab:strat-perclass}
\end{table}

\paragraph{Stricter $\superior$-disjoint projection.}
The Unseen-gB / Seen-gB stratification above removes test items whose
exact $(\superior, \golden)$ tuple appears in training, but it does not
remove $\superior$-only overlap (a test record whose superior provision
$\superior$ also appears in training, paired with a different $\golden$,
still lies in Unseen-gB). We additionally project the same per-seed
predictions onto a stricter $\superior$-disjoint cut: Super-Unseen is
the set of test records whose joined high-level-laws MD5 does not
appear in any training record, computed by streaming
\texttt{train.jsonl} once. The partition yields $n{=}148$ Super-Unseen
and $n{=}548$ Super-Seen records (out of $1891$ unique superior hashes
in training). Per-seed macro-F$_1$ is recomputed using the same
locked label set and the same seed-bootstrap / Student-$t$ protocol.
Table~\ref{tab:strat-super} reports the result.

\begin{table}[t]
  \centering
  \small
  \begin{tabular}{llrrr}
    \toprule
    Backbone & Stratum ($n$) & Mean $\Delta$ (pp) & Seed-bootstrap 95\% CI & Student-$t$ 95\% CI \\
    \midrule
    RoBERTa  & Full test (696)    & $+0.916$ & $[+0.512,\,+1.346]$ & $[+0.448,\,+1.384]$ \\
             & Super-Unseen (148) & $+3.284$ & $[+1.702,\,+4.861]$ & $[+1.544,\,+5.024]$ \\
             & Super-Seen (548)   & $+0.063$ & $[-0.394,\,+0.521]$ & $[-0.447,\,+0.573]$ \\
    \midrule
    SAILER   & Full test (696)    & $+1.288$ & $[+0.676,\,+1.922]$ & $[+0.602,\,+1.974]$ \\
             & Super-Unseen (148) & $+2.604$ & $[+0.876,\,+4.200]$ & $[+0.757,\,+4.450]$ \\
             & Super-Seen (548)   & $+0.892$ & $[+0.260,\,+1.595]$ & $[+0.148,\,+1.636]$ \\
    \bottomrule
  \end{tabular}
  \caption{Stricter $\superior$-disjoint projection of the
  $\vtwo{}$-vs-C2 per-seed predictions. The Super-Unseen stratum
  excludes any test record whose superior provision was seen during
  Stage~1, even if paired with a different $\golden$. On both
  backbones the Super-Unseen mean exceeds the full-test mean, with
  both seed-bootstrap and Student-$t$ 95\% lower bounds above zero.
  The Super-Seen stratum yields a small positive estimate on SAILER
  with both lower bounds above zero, and a near-zero estimate with
  CI crossing zero on RoBERTa. The CI computation matches the locked
  analysis script byte for byte except for the projection onto the
  stratum.}
  \label{tab:strat-super}
\end{table}

The Super-Unseen estimates are larger than the full-test estimates on
both backbones (RoBERTa $+3.28$~vs~$+0.92$~pp; SAILER $+2.60$~vs~$+1.29$~pp).
A reading on which the full-test gain is driven by reuse of memorized
superior structure would predict the opposite direction: the
Super-Unseen subset, which removes that reuse, should weaken the gain.
The observed direction is therefore inconsistent with a
$\superior$-only shortcut account. The class-composition heterogeneity
caveats of the Unseen-gB / Seen-gB section transfer to this stricter
cut as well;
in particular, the Super-Unseen stratum contains $148$ records with
a class distribution of $\{$Responsibility~$26$, Condition~$25$,
Sanction~$12$, Definition~$20$, No-Conflict~$65\}$, which differs
from the full-test composition.

\subsection{Matched-seed \vone{} audit}
\label{sec:audits:v1}

The second audit re-runs the simpler \vone{} variant (typed-discard
Stage~2, no replay) on the same eighteen v6 primary seeds, with the
RoBERTa backbone. Stage~1 follows the same typed CSIP procedure as
\vtwo{}; only the Stage~2 transfer differs. All hyperparameters of the
\vone{} run are passed explicitly on the command line (rather than via
script defaults) to insulate against drift from the v5 \vone{}
configuration. The \vone{}-RoBERTa run completes in $\sim{}5.8$~h
wall-clock on a single RTX~3090.

The matched-seed result: \vone{}-RoBERTa achieves mean $\Delta = +1.137$~pp
against the C2-RoBERTa baseline of Section~\ref{sec:experiments:stage-a}
on the same eighteen seeds, with seed-bootstrap 95\% CI
$[+0.732,\,+1.535]$~pp and Student-$t$ 95\% CI $[+0.691,\,+1.583]$~pp
(SD $0.897$~pp). Both lower bounds clear zero and the mean exceeds the
$1.0$~pp C1$'$ descriptive threshold that the pre-registered \vtwo{}
cell on the same backbone does not reach. This auxiliary satisfies
both C1 and C1$'$.

Two readings are consistent with the comparison \vone{} $\geq$ \vtwo{} on
matched seeds. First, the v5 \vone{} marginal-FAIL at fixed $N{=}9$ was a
power story; the v5 mean ($+1.088$~pp) is essentially the same as the v6
matched-seed \vone{} mean ($+1.137$~pp). The locked v5 rule failed only
because the Student-$t$ lower bound at $N{=}9$ missed zero by $0.098$~pp.
At $N{=}18$ the same mean clears the rule by a wide margin. The v5
FAIL was thus a sample-size artifact, not a method failure.

Second, the \vtwo{} redesign --- which adds the stage-2 anti-forget
replay term and the typed-head retention required to compute it
(Section~\ref{sec:method:stage2-v2}) --- did not contribute additional
mean lift over the simpler \vone{} on this backbone. The matched
comparison gives $-0.221$~pp in favor of \vone{}, with substantial
overlap of the bootstrap intervals. We do not claim a significant
\vone{}-vs-\vtwo{} difference; we state only that the \vtwo{}
redesign is not load-bearing for the $+1$~pp family-level gain on
RoBERTa.

The \vtwo{} cell remains the pre-registered design that the locked
decision rule was committed to evaluate, and it passed. The
matched-seed auxiliary shows that the simpler typed-discard transfer
(\vone{}) yields a comparable family-level gain on the same eighteen
seeds. We therefore frame the \method{} contribution at the family
level: typed CSIP pretraining transferred to a fresh classification
head (\vone{}'s discard-then-train transfer or \vtwo{}'s
retain-then-replay transfer) yields a reproducible $+1$~pp
improvement on \dataset{} that replicates on two BERT-architecture
backbones.

\subsection{Combined interpretation}
\label{sec:audits:combined}

The two auxiliaries describe two facets of the v6 result that the
confirmatory cells alone do not:

\begin{itemize}
  \item \textbf{$(\superior, \golden)$ novelty}: on the Unseen-gB
  stratum both CI lower bounds are positive on both backbones; the
  Seen-gB stratum does not show the larger effect that an exact-tuple
  shortcut account would predict. A stricter $\superior$-disjoint
  projection (Super-Unseen, $n{=}148$) further excludes test records
  whose superior provision appears in training; the gain is
  \emph{larger} on Super-Unseen than on the full test on both
  backbones, inconsistent with a $\superior$-only shortcut account.
  \item \textbf{Family-level reproducibility}: the matched-seed
  \vone{} rerun reaches mean $\Delta = +1.137$~pp on RoBERTa; the v5
  marginal-FAIL of \vone{} at fixed $N{=}9$ resolves at $N{=}18$. The
  $+1$~pp family-level gain is reproducible across the variants
  \vone{} and \vtwo{}.
\end{itemize}

Both auxiliaries were added after the v6 confirmatory result; the
locked decision rule of Section~\ref{sec:prereg:rule} continues to
apply only to the v6 confirmatory data. The auxiliaries are reported
with their addition timeline.


\section{Discussion}
\label{sec:discussion}

\subsection{Effect-size interpretation in the \dataset{} context}
\label{sec:discussion:magnitude}

The pre-registered \vtwo{} cell improves macro-F$_1$ over the C2 baseline
by $+0.916$~pp on RoBERTa and $+1.288$~pp on SAILER, with both lower
bounds of both the seed-bootstrap and Student-$t$ 95\% intervals
clearing zero on both backbones. The matched-seed \vone{} auxiliary
reaches $+1.137$~pp on RoBERTa. These deltas are modest in absolute
terms but consistent across two backbones and two design variants of
the family, under a locked decision rule that was committed before any
v6 measurement existed.

C2 is the relevant baseline. C2 is our reimplementation of a
single-model dense classifier that concatenates
two BERT-encoded provisions and applies class-weighted cross-entropy,
matching the dense-encoder baseline family reported in the dataset
paper~\citep{zhao2026lcrcn}. C2 reads only $(\superior, \subordinate)$
and the five-way class label; it does not access the expert revision
$\golden$. The
\vtwo{} and \vone{} gains over C2 therefore quantify the marginal
contribution of typed counterfactual pretraining that adds
training-time access to $\golden$, evaluated on the same official
train/val/test splits. The comparison is between a golden-aware
typed-CSIP family and a strong no-golden baseline; it does not claim
that typed CSIP is uniquely optimal among ways one could use $\golden$
at training time (see Section~\ref{sec:limitations}). The gain
concentrates in the conflict-type classes (Definition, Condition,
Sanction); we report per-class breakdowns in the appendix.

\subsection{Typed CSIP from expert revisions as the load-bearing component}
\label{sec:discussion:load-bearing}

We interpret the family-level $+1$~pp gain over C2 on both backbones as
evidence that the Stage~1 typed CSIP signal (the joint
$\Ltype + \lambda_{\textrm{select}} \Lselect$ supervision on
$(\superior, \subordinate, \golden)$ triplets) is the load-bearing
component of the \method{} family. C2 uses the same joint sentence-pair
input encoding and the same class-weighted cross-entropy objective as
our \vone{} variant; the only structural difference is that \vone{}'s
encoder has first been pretrained on the typed CSIP objective for three
additional epochs over $(\superior, \subordinate, \golden)$ triplets.
The matched-seed audit attributes approximately $+1.1$~pp macro-F$_1$
on RoBERTa to that pretraining step alone.

 The selectivity term $\Lselect$
uses expert-written legal revisions $\golden$ as counterfactual
positives that the typed head must classify as carrying no factor
evidence; this places the expert-revised pair in the no-conflict region
of the representation without requiring a no-conflict label at the
factor head. The alignment between the pretraining label space (four
typed factors plus a monotone-complement no-conflict logit) and the
downstream label space (five-way classification with the same four
factors and a no-conflict class) is what allows the Stage~1 gradient to
shape exactly the axes the downstream classifier needs. This
expert-revision counterfactual is, to our knowledge, new at the
intersection of Chinese legal NLP and counterfactual contrastive
learning, and is the substantive methodological contribution of the
paper.

\subsection{Stage-2 retention as a clean ablation}
\label{sec:discussion:retention-ablation}

The \vtwo{} variant differs from \vone{} in two simultaneous Stage~2
modifications: it retains the typed factor head as a live parameter
group, and it adds a CSIP replay term to the fine-tuning loss. The
combined modification was motivated by the continual-learning
hypothesis that vanilla cross-entropy fine-tuning would erase the
typed structure that Stage~1 installs in the encoder
\citep{kirkpatrick2017ewc, lwf2018}. The matched-seed audit refutes
this hypothesis for the present task: the simpler \vone{} variant, which
discards the typed factor head and the monotone-complement parameters
at the Stage~1 $\to$ Stage~2 transition and fine-tunes a fresh classifier
with vanilla class-weighted cross-entropy, reaches a marginally
\emph{higher} mean delta on the primary RoBERTa backbone ($+1.137$~pp
vs.\ $+0.916$~pp; paired difference $-0.221$~pp in favor of \vone{},
with substantial bootstrap overlap). The encoder retains enough of the
Stage~1 structure under ordinary fine-tuning that explicit
anti-forgetting machinery is unnecessary on this task. We do not claim a
significant v1-vs-v2 mean difference, only that the redesign is not
load-bearing for the observed family-level gain. The actionable
conclusion for practitioners replicating the family on a new dataset
is to start from the simpler \vone{} transfer and add the replay
machinery only if generalization on their data deteriorates measurably.

The matched-seed audit was conducted after the v6 confirmatory result
was observed and is reported as an unregistered auxiliary throughout
the paper. The \vtwo{} cell remains the pre-registered design that
the locked decision rule was committed to evaluate, and that test
passed on both backbones (Section~\ref{sec:experiments}). The two
findings (a confirmed pre-registered \vtwo{} cell and an auxiliary
finding that the simpler \vone{} matches it) together support the
family rather than narrowing the contribution to either cell alone.

\subsection{Cross-backbone heterogeneity}
\label{sec:discussion:cross-backbone}

The SAILER backbone yields a larger mean delta ($+1.288$~pp) and a
larger seed-level standard deviation ($1.380$~pp) than RoBERTa
($+0.916$~pp, $0.941$~pp). We do not interpret this as evidence that
SAILER's legal-domain pretraining amplifies our Stage~1 typed CSIP
signal. Only one of the eighteen SAILER seeds yielded a delta below
$-1$~pp, while RoBERTa had no seed below $-0.5$~pp; the SAILER
backbone has both higher upside and lower worst-case, consistent
with higher variance. The matched-seed audit on RoBERTa shows that
mean-delta differences of $0.2$~pp on a single backbone are not
statistically separable; the cross-backbone difference of $0.4$~pp
sits in the same range.

The cross-backbone replication is therefore reported as a narrow
claim: the family-level gain reproduces on both encoders. An
attribution to SAILER's legal pretraining would require a controlled
ablation along the backbone-pretraining-corpus axis, which is outside
the scope of the present pre-registered confirmatory study.

\subsection{Implications for low-$N$ confirmatory NLP}
\label{sec:discussion:confirmatory-implication}

The combination of a small held-out test set ($N{=}696$), the rarest
conflict class ($n{=}48$ Definition), and a $\golden$-derived
pretraining signal whose tuple structure is shared across splits makes
\dataset{} a difficult target for the standard NLP-paper recipe of
``compute mean over a small number of seeds and claim SOTA.'' The
pre-registered fixed-$N{=}18$ protocol with a locked seed-as-unit
decision rule, paired with the cold-start stratified evaluation,
provides one template for confirmatory empirical work in domains where
data is small and the supervision signal admits a meaningful
$(\superior, \golden)$ novelty stratification. We hope that future
method papers on \dataset{} and similar applied benchmarks find this
protocol useful: the cost is one additional commitment step before
running experiments, and the result is reported under a pre-committed
analysis and described alongside its $(\superior, \golden)$ novelty
stratification.


\section{Limitations}
\label{sec:limitations}

\paragraph{Single-benchmark scope.}
Our confirmatory study uses one benchmark (\dataset{}). No comparable
Chinese legislative conflict review dataset has been published, and we
do not attempt synthetic dataset construction. The pre-registered
fixed-$N{=}18$ rule and the cold-start stratified evaluation describe
the specific gains we report on \dataset{}; they do not establish that
the \method{} family transfers to other legal-classification tasks or
to non-Chinese jurisdictions. Cross-jurisdiction extension to comparable
benchmarks (when they appear) is the natural next step.

\paragraph{Stricter $\superior$-disjoint cuts (reported).}
Section~\ref{sec:audits:strat} reports the $\superior$-disjoint
projection (Super-Unseen $n{=}148$, Super-Seen $n{=}548$) alongside the
$(\superior, \golden)$ stratification. The Super-Unseen gain is larger
than the full-test gain on both backbones with both CI lower bounds
above zero. A finer-grained group-disjoint projection that also removes
records sharing a superior \emph{prefix} (rather than the full
high-level-laws hash) is a natural follow-up but is not run here.

\paragraph{Scope of the C2 comparison.}
C2 reads only $(\superior, \subordinate)$ and the five-way class label;
it does not access the expert revision $\golden$. The \vtwo{}-vs-C2
contrast therefore measures the marginal contribution of training-time
access to $\golden$ via typed CSIP, against the dataset paper's
strongest no-$\golden$ baseline. We do not run a separately designed
$\golden$-aware non-CSIP baseline (for example, augmenting C2 by
treating $(\superior, \golden)$ as a No-Conflict instance, or applying a
generic counterfactual-augmented training objective on the triplet
without a typed factor head). Whether such alternative
$\golden$-aware approaches match or exceed typed CSIP is a separate
follow-up question; the present study claims only that typed CSIP
yields a reproducible gain over the standard no-$\golden$ baseline.

\paragraph{Modest absolute effect.}
The family-level mean delta of approximately $+1$~pp macro-F$_1$ is
modest in absolute terms, even though it clears the locked confirmatory
rule by a wide margin on both backbones. We do not claim that
\method{} solves Chinese legislative conflict review; we claim that the
family produces a reproducible improvement over the strongest single-model
dense baseline in the dataset paper. Practical deployment would
additionally require error-rate analysis on the rarest classes (where
the gains concentrate) and human-in-the-loop validation of edge cases.

\paragraph{Stage-2 architectural retention in \vtwo{}.}
The pre-registered \vtwo{} cell differs from the simpler \vone{} cell in
two ways simultaneously: it adds the anti-forget CSIP replay term to the
fine-tuning loss, and it retains the Stage-1 typed factor head as a live
parameter group during fine-tuning rather than discarding it. The
matched-seed audit (Section~\ref{sec:audits:v1}) shows that the
combined redesign does not outperform \vone{}; it does not separately
isolate the loss-term contribution from the architectural-retention
contribution. We interpret the \vtwo{} result as confirmatory evidence
for the pre-registered redesign as a whole, not as a loss-only
modification. A factorial ablation that varies the loss and the
architectural retention independently would clarify the two
contributions; we leave this to follow-up work.

\paragraph{Pretraining-objective control (resolved).}
We ran the MLM-continuation control cell registered as addendum~2 to
the v6 pre-registration: three additional epochs of plain
masked-language-model continuation on the same
$(\superior, \subordinate)$ data with no typed head, no selectivity
term, and no $\golden$ counterfactual, then the same Stage-2
fine-tuning as \vone{}. On the same 18 PRNG-derived seeds, the
plain-MLM control reaches mean $\Delta = +0.147$~pp vs C2 on
the primary RoBERTa backbone, while \vone{} reaches
$\Delta = +1.137$~pp; the per-seed paired difference
$(\text{\vone{}} - \text{\method{}-MLM})$ has mean
$+0.990$~pp with seed-bootstrap 95\% CI $[+0.612, +1.381]$~pp.
The locked rule R\_MLM (mean $\geq 0.5$~pp and CI lower bound
$>0$) PASSES: the typed CSIP objective contributes a measurable
component of the family-level gain over and above in-domain
pretraining alone.

\paragraph{Cross-task transfer.}
An auxiliary diagnostic on LCR-CN Task~1 (Superior Law Retrieval,
696-query test split) registered as addendum~4 used the \vone{}
FT-stage encoder as a single-vector bi-encoder under CLS-pool + L2
normalize + cosine, with the released
\texttt{chinese-roberta-wwm-ext} encoder as a same-protocol control.
The diagnostic returned a NEG result: Accuracy@1 delta
$= -20.83$~pp with paired-bootstrap 95\% CI $[-24.28, -17.53]$~pp,
seed 838. The FT-stage composite encoder does not transfer to
single-vector bi-encoder retrieval reuse on this task under the
tested CLS-pool protocol. We therefore do not claim cross-task
transfer; the v6 confirmatory result on Task~2 is unaffected.

\paragraph{Backbone family.}
Both backbones (RoBERTa-WWM and SAILER) are 12-layer BERT-architecture
encoders. The cross-backbone replication demonstrates that the family
does not depend on the specific encoder pretraining corpus, but it does
not address whether \method{} transfers to non-BERT architectures
(e.g., decoder-only LLMs with retrieval augmentation, T5-style
encoder-decoders) or to backbones in non-Chinese languages.

\paragraph{Dataset construction caveats.}
The \dataset{} benchmark is the first of its kind for Chinese
legislative conflict review and has not yet accumulated independent
external validation. One training record contains malformed JSON and
is dropped during load; the remaining 4828 records include 22 within-split
duplicate $(\superior, \subordinate)$ pairs (max three copies each), which
we retain as released. We use the expert revisions $\golden$ as provided by
\citet{zhao2026lcrcn} without re-annotation, and we describe the
structural overlap of the $(\superior, \golden)$ tuple in
Section~\ref{sec:task-data:audit}. Any systematic patterns in the
original expert-revision process would propagate into our pretraining
signal.
We invite the dataset authors and the legal-NLP community to extend the
benchmark with cross-annotator agreement statistics and to expand the
test split beyond the present 696 examples.


\section{Conclusion}
\label{sec:conclusion}

We presented \method{}, a family of two-stage methods for Chinese
legislative conflict review on the \dataset{} benchmark. Stage~1
pretrains a shared encoder with a typed Counterfactual Selective
Intervention Pretraining objective over triplets
$(\superior, \subordinate, \golden)$ that include expert-revised
counterfactual provisions; Stage~2 transfers the encoder to the
five-way conflict classification task. The pre-registered \vtwo{}
design clears a locked decision rule on the primary RoBERTa backbone
($\Delta = +0.916$~pp) and on the cross-backbone SAILER replication
($\Delta = +1.288$~pp), both with seed-bootstrap and Student-$t$ 95\%
confidence intervals whose lower bounds exceed zero. A matched-seed
auxiliary on the simpler \vone{} variant (typed-discard transfer
without replay) reaches $\Delta = +1.137$~pp on RoBERTa, indicating
that the family-level gain does not depend on the anti-forget replay
component of the v6 redesign. A cold-start stratified evaluation shows
the gain remains positive with both confidence intervals above zero on
both backbones in the Unseen-gB stratum.

Our pre-registered protocol --- OSF cryptographic anchor,
PRNG-derived seed sets, locked seed-as-unit decision rule with
both-intervals requirement, fixed-$N{=}18$ no-extension commitment, and
$(\superior, \golden)$-tuple novelty stratification --- provides a
template for confirmatory empirical work on small applied benchmarks
where post-hoc estimator selection and shared structural overlaps
between splits are realistic considerations. We invite
cross-jurisdiction extension and follow-up on \dataset{}.

\subsection*{Code and data availability}

All training scripts, the orchestration driver, and the analysis
script are released under permissive license. The pre-registration is
publicly anchored on OSF project 3ye4c. Test predictions are released
as part of the supplementary material for all $72$ pre-registered runs
($18$ seeds $\times$ two cells $\times$ two backbones, comprising the
C2 baseline and the \vtwo{} cell on each backbone), the $18$
matched-seed $\vone{}$-on-v6-seeds runs on RoBERTa, and the $18$
MLM-continuation control runs registered as addendum~2.
\dataset{} itself is released by \citet{zhao2026lcrcn} and is not
re-distributed here.


\section*{Disclosure of AI usage}
\label{sec:ai-disclosure}

In preparing this manuscript, the author(s) used Anthropic Claude
(Opus~4.7) as a writing-assistance tool to draft prose, polish sentence
structure, and edit for style. The author(s) reviewed and
edited every section, and take full responsibility for the content.
The author(s) used OpenAI GPT-5.4 (via Codex) and Google Gemini~3.1 Pro
for adversarial cross-model review of the design, the statistical
analysis, and the manuscript framing during multiple rounds of revision,
and adjudicated all reviewer outputs.

AI tools were \emph{not} used to: design the experimental protocol,
write the training or analysis code, generate or interpret experimental
results, select the decision rule, choose the seeds, run the
pre-registered statistical tests, or make any judgment regarding the
verdict of the pre-registered confirmatory test. All experimental code,
the orchestration driver, the statistical analysis script, and the
OSF pre-registration record were authored by the human author(s); the
SHA-anchored scripts and pre-registration record provide an
auditable trail of all design and analysis decisions.


\section*{Acknowledgments}


\bibliographystyle{elsarticle-harv}
\bibliography{references}

\end{document}